\documentclass[conference]{IEEEtran}
\IEEEoverridecommandlockouts
\usepackage{cite}
\usepackage{amsmath,amssymb,amsfonts}
\usepackage{algorithm}
\usepackage{algorithmic}
\usepackage{graphicx}
\usepackage{textcomp}
\usepackage{xcolor}
\usepackage{multirow}
\usepackage{boldline}
\usepackage{hyperref}
\usepackage{caption}
\usepackage{subcaption} 
\usepackage{multirow}
\usepackage{colortbl}
\def\BibTeX{{\rm B\kern-.05em{\sc i\kern-.025em b}\kern-.08em
    T\kern-.1667em\lower.7ex\hbox{E}\kern-.125emX}}

\makeatletter
\def\@fnsymbol#1{\ifcase#1\or 1\or 2\or 3\or 4\or 5\else\@ctrerr\fi}
\makeatother

\begin{document}

\title{Is Complete Labeling Necessary? Understanding Active Learning in Longitudinal Medical Imaging\\
}
% \author{\IEEEauthorblockN{Siteng Ma\IEEEauthorrefmark{1}\IEEEauthorrefmark{2},
% Honghui Du\IEEEauthorrefmark{1}\IEEEauthorrefmark{2} Prateek Mathur\IEEEauthorrefmark{1}\IEEEauthorrefmark{2}, 
% Brendan S. Kelly\IEEEauthorrefmark{2}\IEEEauthorrefmark{5}, Ronan P. Killeen\IEEEauthorrefmark{3}\IEEEauthorrefmark{4}, Aonghus Lawlor\IEEEauthorrefmark{1}\IEEEauthorrefmark{2}, Ruihai Dong\IEEEauthorrefmark{1}\IEEEauthorrefmark{2}}
% \IEEEauthorblockA{\IEEEauthorrefmark{1}School of Computer Science, University College Dublin, Ireland \\
% Email: \{siteng.ma,honghui.du,prateek.mathur,brendan.kelly,aonghus.lawlor,ruihai.dong\}@insight-centre.org}
% \IEEEauthorblockA{\IEEEauthorrefmark{2}Insight Centre for Data Analytics, Dublin, Ireland}
% \IEEEauthorblockA{\IEEEauthorrefmark{3}School of Medicine, University College Dublin, Dublin, Ireland}
% \IEEEauthorblockA{\IEEEauthorrefmark{4}Department of Radiology, St Vincent's University Hospital, Dublin, Ireland\\
% Email: ronan.killeen@st-vincents.ie}
% \IEEEauthorblockA{\IEEEauthorrefmark{5}Great Ormond Street Hospital for Children, London, United Kingdom}}

\author{
Siteng Ma\textsuperscript{1,2},
Honghui Du\textsuperscript{1,2},
Prateek Mathur\textsuperscript{1,2},
Brendan S. Kelly\textsuperscript{2,5},\\
Ronan P. Killeen\textsuperscript{3,4},
Aonghus Lawlor\textsuperscript{1,2},
Ruihai Dong\textsuperscript{1,2} \\
\textsuperscript{1}School of Computer Science, University College Dublin, Ireland \\
\textsuperscript{2}Insight Centre for Data Analytics, Dublin, Ireland \\
\textsuperscript{3}School of Medicine, University College Dublin, Dublin, Ireland \\
\textsuperscript{4}Department of Radiology, St Vincent's University Hospital, Dublin, Ireland \\
\textsuperscript{5}Great Ormond Street Hospital for Children, London, United Kingdom \\
Emails: {\{siteng.ma,honghui.du,prateek.mathur,aonghus.lawlor,ruihai.dong\}@insight-centre.org},\\
{brendan.kelly@insight-centre.org}, {ronan.killeen@st-vincents.ie}
}

\maketitle

% \author{\IEEEauthorblockN{1\textsuperscript{st} Siteng Ma}
% \IEEEauthorblockA{\textit{Insight Centre for Data Analytics, School of Computer Science, University College Dublin, Ireland} \\
% \and
% \and
% \and
% \and
% \and

\maketitle

\begin{abstract}
Detecting changes in longitudinal medical imaging using deep learning requires a substantial amount of accurately labeled data. However, labeling these images is notably more costly and time-consuming than labeling other image types, as it requires labeling across various time points, where new lesions can be minor, and subtle changes are easily missed. Deep Active Learning (DAL) has shown promise in minimizing labeling costs by selectively querying the most informative samples, but existing studies have primarily focused on static tasks like classification and segmentation. Consequently, the conventional DAL approach cannot be directly applied to change detection tasks, which involve identifying subtle differences across multiple images. In this study, we propose a novel DAL framework, named Longitudinal Medical Imaging Active Learning (LMI-AL), tailored specifically for longitudinal medical imaging. By pairing and differencing all 2D slices from baseline and follow-up 3D images, LMI-AL iteratively selects the most informative pairs for labeling using DAL, training a deep learning model with minimal manual annotation. Experimental results demonstrate that, with less than 8\% of the data labeled, LMI-AL can achieve performance comparable to models trained on fully labeled datasets. We also provide a detailed analysis of the method’s performance, as guidance for future research. The code is publicly available at https://github.com/HelenMa9998/Longitudinal\_AL.

% https://github.com/HelenMa9998/Longitudinal\_AL. 
\end{abstract}


\section{Introduction}
A longitudinal study in medical image analysis involves tracking and analysing a series of medical images over time to observe and measure subject-specific changes in a patient’s condition. This plays a pivotal role in managing chronic and progressive diseases (e.g., tracking tumor growth and shrinkage, assessing the progression of Multiple Sclerosis) \cite{zhao2021longitudinal}. 

Recently, supervised deep learning (DL) approaches that train the deep model using a fully labeled dataset have become well known for effectively identifying and extracting meaningful information in computer vision and medical image analysis \cite{cui2019rnn}. However, their effectiveness is heavily dependent on the availability of extensive datasets with accurate labels \cite{cheplygina2019not}, which poses a major challenge in the context of longitudinal medical imaging \cite{zhao2021longitudinal}. Labeling longitudinal medical images is notably more laborious, costly, and time-consuming \cite{dufresne2020joint} compared to other scenarios. It requires experienced experts to label the data across multiple time points \cite{martin2021current}. Moreover, given that the changes in the images can be minute (3-10 voxels in size on MRI images) and may present with subtle appearances \cite{sepahvand2020cnn}, it is easy to fail in annotating minor abnormalities or changes, leading to inaccuracies and inconsistencies in labeling \cite{sepahvand2020cnn,gerig2016longitudinal}. Therefore, this demands experts label the data very carefully and sometimes repeatedly to ensure all possible changes are annotated, making the process of labeling large longitudinal medical image datasets impractical. 

Deep Active Learning (DAL) is a machine learning approach where the algorithm selectively queries the most valuable data examples for training deep learning models. Oracles (e.g., radiologists) label these examples to minimize the overall labeling effort and maintain or enhance the model’s predictive performance \cite{settles2009active}. Recent research \cite{ozdemir2021active,shen2021labeling,nath2022warm,ma2024adaptive} has demonstrated that DAL can effectively reduce labeling costs in 2D/3D medical image analysis. 

As noted, labeling and training with longitudinal medical images can be much more challenging than with standard medical images. However, no studies have been conducted to explore the effectiveness of DAL with longitudinal medical imaging data. Consequently, it remains unclear what an effective query strategy would be for selecting images in pairs to train a DL model for change detection effectively. Simply applying existing DAL methods to longitudinal medical images may result in poor performance. Therefore, this paper aims to answer the following research question: \textit{How can DAL be used to reduce the labeling effort in longitudinal medical imaging while maintaining the model's performance? }

To answer this question, we propose a novel approach named Longitudinal Medical Imaging Active Learning (LMI-AL). LMI-AL is the first framework specifically designed to adapt DAL approaches to change detection in longitudinal medical imaging, significantly reducing labeling efforts by enhancing the efficiency of DL model training. Initially, LMI-AL transforms 3D images into 2D slices. Each pair of slices (baseline and follow-up scans) and their differences are grouped together. To ensure the model learns from all potential variations, LMI-AL creates the initial slice pool using all possible pairwise combinations of slices. LMI-AL then iteratively selects slice pairs and their corresponding differences based on the selected AL query strategy for labeling and trains the DL model accordingly. Through extensive experiments, we demonstrate that LMI-AL achieves comparable performance to fully supervised methods while requiring less than 8\% of labels, significantly decreasing the annotated image slices. We also provide detailed analysis based on the result to give guidance for future researchers.

The rest of the paper is organized as follows: we present a survey of related research in Section \ref{Related Work}, the proposed LMI-AL framework is detailed in Section \ref{Proposed Method}. Then Section \ref{Experiment Setup} and Section \ref{Experiment Results} present the experiment setup, results, and corresponding analysis of our empirical studies. The conclusion can be found in Section \ref{Conclusion}.

\begin{figure*}[tb]
\centering
\includegraphics[scale = 0.7]{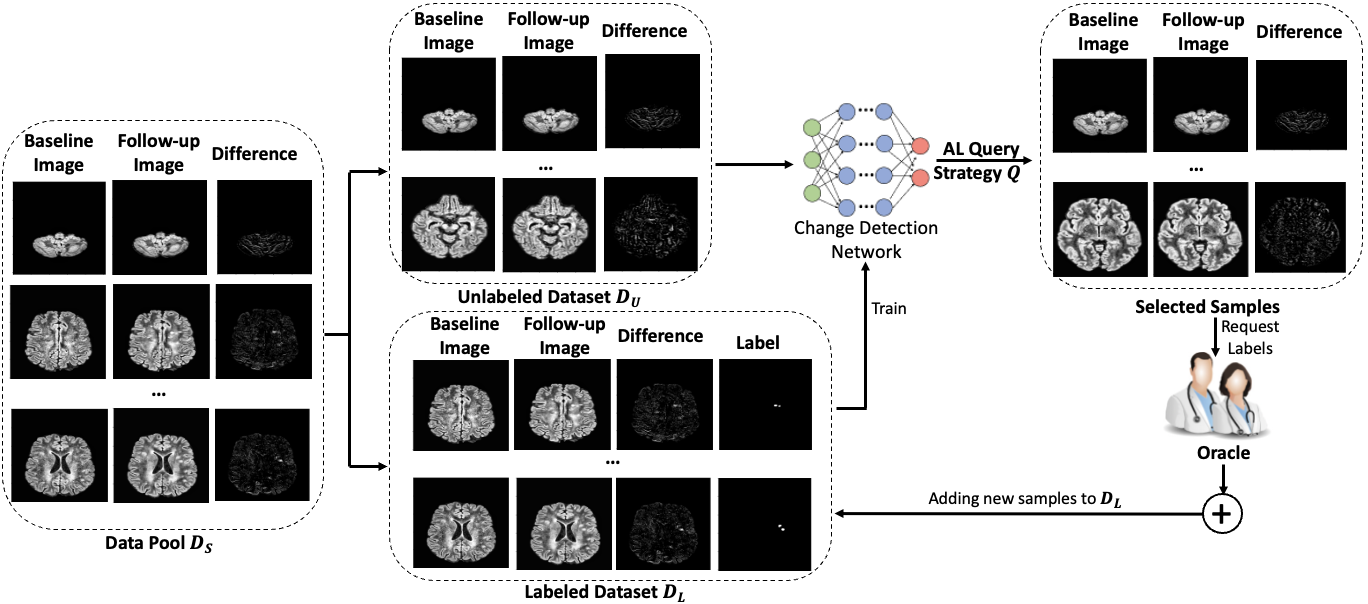}
\caption{The overall process of applying active learning in longitudinal medical image analysis. }
\label{fig1}
\end{figure*}

\section{Related Work}
\label{Related Work}
In this section, we present an overview of DL change detection models and methods in longitudinal medical settings. Then Deep Active Learning (DAL) methods in general scenarios are given.

\subsection{Change Detection in Longitudinal Medical Imaging}
A majority of literatures \cite{cui2019rnn,he2023longitudinal,basaran2022new} have focused on leveraging fully supervised learning for solving problems in longitudinal medical image analysis. 

Initially, Recurrent Neural Network (RNN)-based models (e.g., LSTM \cite{hochreiter1997long} and GRU \cite{gulcehre2014learned}) are extensively used to capture temporal relationships in longitudinal medical imaging data. These approaches typically involve using convolutional neural networks (CNNs) to extract features from images, which are then processed by RNN-based models to learn temporal dependencies \cite{cui2019rnn,xu2019deep}. However, RNN-based models are difficult to train and prone to the exploding gradient problem \cite{ribeiro2020beyond}. Recent studies have integrated LSTM layers into CNNs to better model temporal images \cite{dixit2021assessing,he2023longitudinal}. 

% The MSSEG-2 challenge \cite{commowick2021msseg} at the 2021 MICCAI conference, focused on detecting new lesions in longitudinal medical imaging, significantly accelerates development and increases interest in this field \cite{kelly2024siamese}. Many studies adopt advanced computer vision techniques to enhance the performance of longitudinal medical imaging in this challenge. Some research \cite{basaran2022new, ashtari2022new, sepahvand2020cnn} concatenates baseline and follow-up images as 2-channel inputs to a UNet model \cite{ronneberger2015u} to enhance performance. In \cite{hitziger2022triplanar}, adjacent temporal sequences across multiple views (e.g., coronal, axial, and sagittal planes) in 3D medical imaging (e.g., MRI or CT scans) are concatenated as UNet inputs, and the outputs are averaged to further improve results.

CNN models have proven effective in image analysis \cite{lecun1989backpropagation}, are frequently used for change detection \cite{basaran2022new, ashtari2022new, sepahvand2020cnn}. For instance, UNet \cite{ronneberger2015u}, a popular CNN-based segmentation architecture, concatenates baseline and follow-up images as 2-channel inputs to enhance model performance \cite{basaran2022new, ashtari2022new, sepahvand2020cnn}, where \cite{sepahvand2020cnn} further involves the difference between baseline and follow-up images to help the model concentrate on the potential change areas. Moreover, UNet remains the core model and employs multiple views, such as coronal, axial, and sagittal planes in 3D medical imaging (e.g., MRI or CT scans) by concatenating adjacent temporal sequences as inputs and averaging the outputs, which can further improve the results \cite{hitziger2022triplanar}. Recent innovations have incorporated attention mechanisms into UNet to better focus the model on target regions \cite{sarica2022new}, maintaining the basic framework described above. 

Due to the typically small size of new lesions, cascaded CNNs \cite{zhu2017face} (e.g., a series of convolutional neural networks are connected sequentially, with the output of one used as the input for the next) are introduced. Researchers use an initial network to identify lesion centers from baseline images, extract patches from both baseline and follow-up images, and then feed these patches into the next model to enhance change detection \cite{salem2022improving}. To handle hard-to-learn cases, \cite{schmidt2022online} adapts Online Hard Example Mining strategy that uses two networks, where the first identifies potential new lesion voxels, and the second reduces misclassified voxels from the first network’s output. Siamese networks \cite{bromley1993signature} are proposed to compare two inputs and determine their similarity, such as face verification and signature verification. Fenneteau et al. \cite{fenneteau2021siamese} utilize this approach to segment each time slot, and then a reﬁnement block concatenates segmentation maps produced for each FLAIR image to predict the ﬁnal segmentation of new lesions. Vision Transformers (ViTs) \cite{dosovitskiy2020image} have gained popularity due to their ability to learn inherent image features and temporal relationships. A recent study \cite{kelly2024siamese} adopts ViTs to enhance learning in longitudinal medical imaging.

Although all these methods have achieved excellent results, they are fully supervised and require large amounts of well-labeled data to train the model, which is impractical in longitudinal medical imaging.

\subsection{Deep Active Learning}
Unlike traditional supervised learning, DAL reduces labeling costs by selectively querying an oracle to label the most informative data points, achieving similar or better performance with fewer labels \cite{settles2009active,ren2021survey}. Every DAL scenario involves determining the information contained in unlabeled instances, which is defined as a query strategy. Uncertainty sampling is the most common query strategy in DAL, where the model selects the most uncertain data points for labeling (e.g., the model’s prediction confidence on an instance is less than 0.5) \cite{settles2009active}. By focusing on labeling uncertain examples, the model reduces overall uncertainty, which improves its ability to make accurate predictions and enhances learning efficiency \cite{lewis1995sequential,scheffer2001active}. Using the softmax output of the deep model \cite{lewis1995sequential} and information entropy \cite{settles2009active,qi2018label} are two common methods for measuring the uncertainty of an instance. Then, as a way to evaluate the uncertainty of the model, \cite{houlsby2011bayesian} introduced the use of Bayesian Convolutional Neural Networks for Active Learning (BALD) to mitigate the issue of the DL model exhibiting excessive confidence in its output results. Monte Carlo (MC) dropout \cite{houlsby2011bayesian} is proposed as an easier way to evaluate Bayesian uncertainty. 

Although the uncertainty-based method is simple in form and has low computational complexity, they consider each sample independently and therefore ignore the correlation between samples \cite{ren2021survey}. Diversity-based query strategy is another major approach in DAL. It selects a diverse set of data points for labeling to cover a broad range of the input space, improving the model’s generalization and reducing redundancy in the training data \cite{sener2018active}. The Coreset technique is proposed as an effective representation learning method to select samples \cite{sener2018active}, by identifying and selecting the subset of data that exhibits substantial deviations from the distribution of the already labeled data. To leverage the advantages of both uncertainty and diversity methods, hybrid methods combine different query strategies to optimize the selection of data points for labeling \cite{citovsky2021batch,li2024hybrid}. 

However, DAL has been typically used for single-image tasks (e.g., classification, segmentation). There is no existing approach that can be directly applied to the medical change detection task. 

\section{Proposed Method}
\label{Proposed Method}
Let $\mathcal{D} = \{ I_{n}^{t} \mid n \in \{1, \dots, N\}, t \in \{1, \dots, T_n\} \}$ denotes a longitudinal medical image dataset that includes images from $N$ patients, where $I_{n}^t \in \mathbb{R}^{h\times w \times c}$ is a 3D image for patient $n$ scanned at a specific time-step $t$; $h,w,c$ represent the height, width, and number of slices of the image, respectively. $T_n \geq 2$ is the total number of images for patient $n$, $T_n$ can vary across patients. Given two unseen images $I_{n'}^t$ as the baseline and $I_{n'}^{t'}$, $t’ > t$ as the follow-up of the same patient $n'$, the goal of change detection in longitudinal medical imaging is to learn a model $f(\cdot)$ from $\mathcal{D}$ that can identify all the new lesions in $I_{n'}^{t’}$ compared to $I_{n'}^t$. Formally, this can be defined as follows:
\begin{equation}
    f(I_{n'}^{t}, I_{n'}^{t'}) = m; m:\mathbb{R}^{h\times w \times c} \rightarrow \{0,1\}
\end{equation}
where $m$ is the binary change map for voxels. The definition can be simplified to 2D settings when $c=1$, so that $I_n^t \in \mathbb{R}^{h\times w}$. In this paper, to better illustrate the proposed method, all $I$ are considered as 3D images.

Annotating 3D images directly can be complex and overwhelming due to the high dimension. It is common to divide 3D images into 2D slices to make the annotation task more manageable \cite{wang2018deepigeos}. Consider that each $I_n^t$ is composed of a set of 2D slices $\mathcal{X}_{n}^t = \{\mathbf{x}_{n,k}^{t}\}_{k=1}^c$, where $\mathbf{x}_{n,k}^{t} \in \mathbb{R}^{h\times w}$ is the $k$th slice of $I_{n}^t$. The dataset $\mathcal{D}$ can be denoted as $\mathcal{D} = \{\mathbf{x}_{n,k}^{t} \mid n \in \{1, \dots, N\}, t \in \{1, \dots, T_n\}, k \in \{1, \cdots, c\} \}$. Let $\mathbf{y}_{n,k}^{t,t'} \in \{0,1\}^{h\times w}$ represents the corresponding 2D change pixel label between a pair of slices $\mathbf{x}_{n,k}^{t'}$ and $\mathbf{x}_{n,k}^{t}$, where $t' > t$. The objective of the proposed method is to minimize the number of labeled slice pairs needed to train $f(\cdot)$ through DAL while maintaining performance similar to that of supervised learning. 

\begin{algorithm}[tb]
\caption{Learning Procedure of LMI-AL}
\label{alg: LMI-AL}
\begin{algorithmic}[1]
\REQUIRE Dataset $\mathcal{D}$, Budget $B$, Deep learning model $f$, Base Active learning approach $Q$, The number of initial query slice pairs $q_0$, The number of querying slice pairs at each iteration $q$.
\STATE Initialize $\mathcal{D}_S$ \label{alg_state: initialization}
\STATE Initialize $\mathcal{D}_L$ by randomly querying $q_0$ slice pairs from $\mathcal{D}_S$ \label{alg_state: initialization_D_L}
\WHILE{$|\mathcal{D}_L| < B$} \label{alg_state: loop_start}
    \STATE $\mathcal{D}_U \leftarrow \mathcal{D}_S - \mathcal{D}_L$ \label{alg_state: update D_U}
    \STATE Train model $f$ on $\mathcal{D}_L$ \label{alg_state: train}
    \STATE $\mathcal{D}_Q \leftarrow$ $Q(\mathcal{D}_U, q, f)$ \label{alg_state: query_end}
    \STATE $\mathcal{D}_L \leftarrow \mathcal{D}_L \cup \mathcal{D}_Q$ \label{alg_state: update D_L}
\ENDWHILE
\end{algorithmic}
\end{algorithm}

The learning pipeline and pseudo-code for LMI-AL are shown in Figure \ref{fig1} and Algorithm \ref{alg: LMI-AL}, respectively. LMI-AL begins dividing the 3D images into 2D slices. Unlike other single-image tasks (e.g., segmentation, classification), change detection requires a pair of images as input (e.g., one baseline and one follow-up slice). To ensure the model can learn from all potential variations and differences between slices and maximize available training information, LMI-AL creates the initial pairwise slice pool $\mathcal{D}_S$ using all possible pairwise combinations of slices. Additionally, due to variations in imaging conditions (e.g., physiological changes, patient movement, and positioning differences), slices of the same location at different times always exhibit slight differences. To better focus on changes, reduce noise, and improve learning efficiency, LMI-AL uses the difference between two slices as an additional input. The initial data pool $D_S$ can be formally written as $\mathcal{D}_S = \{(\mathbf{x}_{n,k}^{t},\mathbf{x}_{n,k}^{t'}, \mathbf{x}_{n,k}^{t'} - \mathbf{x}_{n,k}^{t}) \mid n \in \{1, \dots, N\}, t \in \{1, \dots, T_n - 1\}, t' \in \{2, \dots, T_n\}, t' > t, k \in \{1, \cdots, c\} \}$ (line \ref{alg_state: initialization}).

A small pre-defined number $q_0$ of slice pairs from $\mathcal{D_S}$ are randomly selected and labeled to form the initial labeled pool $\mathcal{D}_L = \{((\mathbf{x}_{n,k}^{t},\mathbf{x}_{n,k}^{t'}, \mathbf{x}_{n,k}^{t'} - \mathbf{x}_{n,k}^{t}), \mathbf{y}_{n,k}^{t,t'}) \mid n \in \{1, \dots, N\}, t \in \{1, \dots, T_n - 1\}, t' \in \{2, \dots, T_n\}, t' > t, k \in \{1, \cdots, c\} \}$ (line \ref{alg_state: initialization_D_L}). Subsequently, LMI-AL commences a loop continues until the size of $\mathcal{D}_L$ reaches a predefined budget $B$ (line \ref{alg_state: loop_start}). In each iteration, the unlabeled set $\mathcal{D}_U$ is updated by excluding the labeled set $\mathcal{D}_L$ from the initial pairwise slice pool $\mathcal{D}_S$ (line \ref{alg_state: update D_U}). The DL model $f$ is then trained on the current labeled set $\mathcal{D}_L$ (line \ref{alg_state: train}) and then used to identify the most informative slice pairs. Based on the DAL strategy $Q$, the model selects $q$ new slice pairs from $\mathcal{D}_U$ to be labeled by an oracle (line \ref{alg_state: query_end}). Any DAL query strategy can be used as $Q$, e.g., uncertainty-based, diversity-based, or hybrid AL methods. These selected pairs are then incorporated into $\mathcal{D}_L$, enriching it with new data for subsequent training iterations (line \ref{alg_state: update D_L}). 

When LMI-AL needs to perform change detection on an unseen pair of 3D images $I_{n',k}^t$ and $I_{n',k}^{t'}$, it first slices both 3D images into 2D sections. For each corresponding pair of 2D slices from the baseline and follow-up images, LMI-AL computes a difference image, with special concern for whitened areas (tumors are white). The original slices and their difference images are then concatenated to form the model input. Each input consists of the baseline slice, the follow-up slice, and the difference image. The model then identifies areas $\mathbf{\hat{y}}_{n',k}^{t,t'}$ where lesions or other changes have occurred between the two time points on each slice. Finally, the model outputs a binary change map $m = \{\mathbf{\hat{y}}_{n',k}^{t,t'} \mid k \in \{1, \cdots, c\}\}$ indicating the detected changes between the 3D images, allowing for detailed analysis of disease progression or treatment response.

\section{Experiment Setup}
\label{Experiment Setup}
\subsection{Datasets}
Multiple Sclerosis (MS) is an autoimmune neuro-degenerative disease that causes inflammation, damaging the myelin sheath around nerves and resulting in evolving lesions and progressive brain atrophy. In clinical practice, this progression is typically monitored through longitudinal brain MRI analysis \cite{dufresne2020joint}. Therefore, in this study, we use two MS longitudinal MRI datasets to evaluate our proposed method.
\begin{itemize}
    \item \textbf{MSSEG-2 Dataset} \cite{commowick2021msseg}: MSSEG-2 is a publicly available dataset consisting of 200 MRI images from 100 MS patients, acquired from 15 different MRI scanners across multiple centers. Each patient was scanned at two time points, typically 1 to 3 years apart. Since not all the data in this dataset are labeled, we use only the MRI images from the 40 patients with available labels in this study.
    \item \textbf{MS Longitudinal MRI Dataset (SVUH)} \cite{kelly2024siamese} \footnote{https://zenodo.org/records/8146426}: This internal dataset was collected from St. Vincent’s University Hospital in Ireland between 2019 and 2022, including only patients who had new lesions. It contains 2467 MRI images from 170 patients, with each patient undergoing up to 5 scanning sessions using 2 different scanners over a period ranging from 16 to 885 days. All data were labeled by two certified radiologists in their first year post-examination using ITK Snap. The labels were validated by a third expert radiologist specializing in neuroradiology.
\end{itemize}
Each dataset contains multiple sequences. Since change detection is similar to segmentation, and the FLAIR sequence is particularly well-suited for visualizing pathology, making it the most commonly used modality for segmentation tasks. Therefore, all the experiments are conducted using the FLAIR sequence. For MSSEG-2, the data are split into 28 patients for training (5,520 slice pairs), 6 for validation (767 slice pairs), and 6 for testing (1,074 slice pairs). For SVUH, the splits are 110 patients for training (29,638 slice pairs), 30 for validation (5,792 slice pairs), and 30 for testing (7,050 slice pairs). To ensure consistency and reliability of the model input, all slices are cropped to the largest brain cross-section, then normalized and resized to 256×256. Data augmentation techniques, including Gaussian blur, flipping, and rotation, are applied to each image in the training dataset.

\subsection{Backbone Model and Active Learning Methods}
% \begin{itemize}
%     \item Least Confidence: Least confidence is a query strategy where the model selects data points with the lowest confidence for labeling. It is measured by the highest softmax probability, targeting areas where the model is most uncertain.
% \end{itemize}
We employ the UNet for biomedical image segmentation \cite{ronneberger2015u} from glass library\footnote{https://github.com/FrancescoSaverioZuppichini/glasses/tree/master} as our backbone model. Given the significant imbalance between the few target pixels and the many background pixels in each slice, focal loss \cite{lin2017focal} ($\alpha = 1, \gamma = 2$) is used to minimize the impact of this imbalance. The UNet is trained for up to 100 epochs with early stopping after 5 epochs and uses the Adam optimizer \cite{kingma2014adam}. To evaluate how effectively DAL reduces labeling effort while maintaining change detection model performance, we compare the model under both DAL and supervised settings. The hyperparameters are tuned separately under supervised and DAL settings on each dataset to obtain optimal performance results. For the supervised setting, the learning rate and batch size are set to 0.0005 and 8, respectively, for both datasets. For AL setting, the learning rate is set to 0.0001 for both datasets, while the batch size is set to 8 for the MSSEG-2 dataset and to 4 for the SVUH dataset, respectively. The number of initial query slice pairs $q_0$ is set to 100 for MSSEG-2 and 500 for SVUH dataset. The number of query slice pairs in each subsequent iteration $q$ is set to 50 for MSSEG-2 and 200 for SVUH.

We select eight well-known and widely used DAL methods, covering uncertainty-based (\textbf{LeastConfidence} \cite{lewis1995sequential}, \textbf{Margin} \cite{scheffer2001active}, \textbf{Entropy} \cite{settles2009active}, \textbf{BALD} \cite{houlsby2011bayesian}), diversity-based (\textbf{Coreset} \cite{sener2017active}), hybrid (\textbf{ClusterMargin} \cite{citovsky2021batch}, \textbf{Hybrid Sampling} \cite{li2024hybrid}), and the common baseline \textbf{Random} as the base AL query strategies $Q$ in our proposed framework to evaluate their performance on longitudinal imaging.

\subsubsection{Least Confidence \cite{lewis1995sequential}} selects samples for which the model has the lowest confidence in its most likely prediction. This is measured by the probability of the most probable label as predicted by the model.

\subsubsection{Margin Sampling \cite{scheffer2001active}} focuses on the difference between the first and second most probable predictions made by the model. Samples are chosen based on the smallest margins, indicating closely competing predictions. 

\subsubsection{Entropy Sampling \cite{settles2009active}} is based on the uncertainty principle, where the selection criterion is the information entropy of the predictive probability distributions of the models. Samples that yield the highest entropy are considered the most uncertain and are thus prioritized for labeling.

\subsubsection{BALD-Dropout \cite{houlsby2011bayesian}} exploits model uncertainty in a Bayesian framework. By utilizing dropout during inference as an approximation of Bayesian posterior, this method selects samples that maximize the mutual information between predictions and model parameters. In our experiment, Dropout is employed with the hyper-parameter n\_drop set to 10.

\subsubsection{K-Center Greedy \cite{sener2017active}} is a core-set approach where the goal is to select samples that are representative of the entire dataset. It chooses samples that are farthest from the current set of labeled samples, effectively covering the input space more uniformly. 

\subsubsection{Cluster Margin Sampling \cite{citovsky2021batch}} combines Hierarchical Agglomerative Clustering (HAC) and margin sampling to enhance the selection process. The dataset is first clustered into several groups, and then margin sampling is applied within each cluster. This method ensures that the samples selected are not only uncertain but representative of the various data clusters, leading to a more balanced dataset for training. It is noteworthy that while this method has been employed for natural image classification tasks, HAC tends to be computationally inefficient when applied to pixel-level segmentation. Consequently, we have replaced HAC with the more conventional KMeans clustering to better suit our scenario. The hyper-parameter n\_clusters and random\_state for KMeans are set to 20 and 42, respectively. Additionally, the first round of selection focuses on diverse samples, choosing n = min(k*10, len(unlabeled\_idxs)), where k is the number of samples selected per round, followed by the uncertainty selection.

\subsubsection{Hybrid Sampling \cite{li2024hybrid}} integrate a Bayesian framework with a hybrid selection method that incorporates density (measured by cosine similarity) and diversity (quantified through mutual information) to select samples that exhibit both high uncertainty, density and diversity. The Bayesian is based on MC Dropout with n\_drop set to 10. Similarly, for the initial step of uncertainty-based method selection, the top 500 most uncertain samples are chosen for subsequent selection based on diversity and density. The influence of diversity compared to density is in a ratio of 2:1.

\subsubsection{Random Selection} involves selecting samples randomly without any heuristic or model-driven strategy. This method serves as a control to assess the efficiency and effectiveness of the other more sophisticated active learning strategies.

\subsection{Performance Metrics}
Due to the similarity between segmentation and change detection, we use common segmentation evaluation metrics (e.g., precision, recall and dice coefficient) to assess the performance of change detection. Precision is the ratio of correctly identified changed voxels to all voxels predicted as changed, indicating the accuracy of the model in detecting true changes. Recall measures the ratio of correctly identified changed voxels to the total actual changed voxels, reflecting the model’s ability to detect all true changes in the 3D data. The dice coefficient evaluates the overlap between the predicted changed voxels and the actual changed voxels, calculated as twice the intersection divided by the sum of the total predicted and actual changed voxels, reflecting the accuracy and completeness of the detection. All experiments are conducted three times, and the average performance across these runs are reported.

% O\begin{tabular}{|l V{5} rrr V{5} c|c|c V{5} c|c|c V{5} c|}

\section{Experiment Results}
\label{Experiment Results}

\subsection{Fully Supervised Learning Result and Analysis}

\begin{table}[htbp]
\centering
\caption{Fully Supervised Learning Performance Evaluation on MSSEG-2 and SVUH Datasets}
\begin{tabular}{|c|c|c|c|c|}
\hline
\textbf{Dataset} & \textbf{Data Used} & \textbf{dice} & \textbf{Rec} & \textbf{Prec} \\
\hline
\multirow{2}{*}{MSSEG-2} & All slices & \textbf{0.4973} & 0.6097 & \textbf{0.4467} \\
\cline{2-5}
& Slices with target & 0.4379 & \textbf{0.6674} & 0.3685 \\
\hline
\multirow{2}{*}{SVUH} & All slices & 0.5907 & 0.1049 & 0.4109 \\
\cline{2-5}
& Slices with target & \textbf{0.6473} & \textbf{0.2654} & \textbf{0.4596} \\
\hline
\end{tabular}
\label{tab1}
\end{table}

Table \ref{tab1} shows the overall supervised learning result for the two datasets. Our experimental evaluation demonstrates distinct performance characteristics when comparing the use of the entirety of available samples versus a focus solely on slices with targets in the MSSEG-2 and SVUH datasets for fully supervised learning. 

For the MSSEG-2 dataset, training with all slices results in better overall metrics compared to using only slices with targets. This can be primarily attributed to the relatively small size of the training set, which necessitates a larger pool of samples to effectively capture the comprehensive distribution of the data. By utilizing all available slices, the model achieves an increase in the overall dice score of 0.4973 compared to 0.4379 with only target-contained slices, and precision increased by 0.08. However, using all slices creates an imbalanced dataset, as many of them do not include the target regions. This imbalance reduces the model's confidence in detecting target areas, as indicated by a drop in recall from 0.6674 to 0.6097.

Conversely, the larger SVUH dataset with more changed cases shows an advantage when focusing solely on slices with targets, which are large enough for model training, substantially reducing the degree of imbalance. This not only boosts the model's confidence in segmenting the target areas but also enhances the recall by 0.16 (0.1049 to 0.2654) and both the precision and dice by approximately 0.05.

\subsection{Active Learning Result}
\begin{table*}[tb]
\caption{Model performance with varied label percentages for the MSSEG-2 (upper section) and SVUH (lower section) datasets. For MSSEG-2, supervised learning achieves 0.4973 (dice), 0.6097 (recall), and 0.4467 (precision). For SVUH, supervised learning results are 0.6473 (dice), 0.2654 (recall), and 0.4596 (precision). The best approach combined with LMI-AL in each column is highlighted in red, while the second-ranked approaches are in bold with a grey background. The methods in bold are the two best methods. The highest dice score is the highest performance achieved during the process, starting after the initial query to 8\% for MSSEG-2 and 5\% for SVUH.}
\begin{center}
\scalebox{0.98}{
\begin{tabular}{|l V{3} c|c|c V{3} c|c|c V{3} c|c|c V{3} c|}
\hline
Percentage of $\mathcal{D}_L$ in MSSEG-2            & \multicolumn{3}{c V{3}}{4\%}                                                                                                                                                                                                                   & \multicolumn{3}{c V{3}}{6\%}                                                                                                                                                                                                                   & \multicolumn{3}{c V{3}}{8\%}                                                                                                                                                                                                                   & \multicolumn{1}{l|}{}                                                       \\ \cline{1-10}
Metrics                    & \multicolumn{1}{l|}{dice}                                                           & \multicolumn{1}{l|}{recall}                                                         & \multicolumn{1}{l V{3}}{precision}                                 & \multicolumn{1}{l|}{dice}                                                           & \multicolumn{1}{l|}{recall}                                                         & \multicolumn{1}{l V{3}}{precision}                                 & \multicolumn{1}{l|}{dice}                                                           & \multicolumn{1}{l|}{recall}                                                         & \multicolumn{1}{l V{3}}{precision}                                 & \multicolumn{1}{l|}{\multirow{-2}{*}{Highest dice}}                         \\ \hline
LMI-AL(Margin)                     & \multicolumn{1}{r|}{0.2637}                                                         & \multicolumn{1}{r|}{0.3048}                                                         & 0.4561                                                         & \multicolumn{1}{r|}{0.3483}                                                         & \multicolumn{1}{r|}{0.4355}                                                         & 0.3981                                                         & \multicolumn{1}{r|}{0.3575}                                                         & \multicolumn{1}{r|}{0.3928}                                                         & 0.4093                                                         & 0.3575                                                              \\  
LMI-AL(LeastConfidence)            & \multicolumn{1}{r|}{0.3471}                                                         & \multicolumn{1}{r|}{0.4114}                                                         & 0.4261                                                         & \multicolumn{1}{r|}{0.3829}                                                         & \multicolumn{1}{r|}{0.5533}                                                         & 0.3710                                                          & \multicolumn{1}{r|}{0.3707}                                                         & \multicolumn{1}{r|}{0.4986}                                                         & 0.3682                                                         & 0.3829                                                                   \\  
LMI-AL(Entropy)                    & \multicolumn{1}{r|}{\cellcolor[gray]{0.85}{\color[HTML]{FE0000} \textbf{0.4537}}} & \multicolumn{1}{r|}{\cellcolor[gray]{0.85}{\color[HTML]{FE0000} \textbf{0.5571}}} & 0.4535                                                         & \multicolumn{1}{r|}{0.4209}                                                         & \multicolumn{1}{r|}{0.4317}                                                         & \cellcolor[gray]{0.85}{\color[HTML]{FE0000} \textbf{0.4897}} & \multicolumn{1}{r|}{0.4357}                                                         & \multicolumn{1}{r|}{0.4797}                                                         & \cellcolor[gray]{0.85}{\color[HTML]{FE0000} \textbf{0.4744}} & 0.4537                                                                       \\ 
LMI-AL(BALD)                       & \multicolumn{1}{r|}{0.3411}                                                         & \multicolumn{1}{r|}{0.3620}                                                          & 0.4233                                                         & \multicolumn{1}{r|}{0.4205}                                                         & \multicolumn{1}{r|}{\cellcolor[gray]{0.85}{\color[HTML]{FE0000} \textbf{0.5633}}} & 0.3892                                                         & \multicolumn{1}{r|}{0.4577}                                                         & \multicolumn{1}{r|}{0.6015}                                                         & 0.3908                                                         & 0.4205                                                               \\ 
LMI-AL(ClusterMargin)              & \multicolumn{1}{r|}{\cellcolor[gray]{0.85}\textbf{0.4132}}                        & \multicolumn{1}{r|}{0.4338}                                                         & \cellcolor[gray]{0.85}{\color[HTML]{FE0000} \textbf{0.5166}} & \multicolumn{1}{r|}{0.3886}                                                         & \multicolumn{1}{r|}{0.4695}                                                         & 0.4079                                                         & \multicolumn{1}{r|}{0.4418}                                                         & \multicolumn{1}{r|}{0.5264}                                                         & \cellcolor[gray]{0.85}\textbf{0.4699}                        & 0.4446                                                            \\ 
LMI-AL(Hybrid)                     & \multicolumn{1}{r|}{0.4027}                                                         & \multicolumn{1}{r|}{0.4735}                                                         & \cellcolor[gray]{0.85}\textbf{0.4889}                        & \multicolumn{1}{r|}{\cellcolor[gray]{0.85}\textbf{0.4215}}                        & \multicolumn{1}{r|}{0.5344}                                                         & 0.4465                                                         & \multicolumn{1}{r|}{0.3963}                                                         & \multicolumn{1}{r|}{\cellcolor[gray]{0.85}{\color[HTML]{000000} \textbf{0.6015}}} & 0.4202                                                         & 0.4620                                                                    \\ 
\textbf{LMI-AL(Random)}            & \multicolumn{1}{r|}{0.4016}                                                         & \multicolumn{1}{r|}{0.4608}                                                         & 0.4663                                                         & \multicolumn{1}{r|}{\cellcolor[gray]{0.85}{\color[HTML]{FE0000} \textbf{0.4279}}} & \multicolumn{1}{r|}{0.5410}                                                          & 0.4418                                                         & \multicolumn{1}{r|}{\cellcolor[gray]{0.85}{\color[HTML]{FE0000} \textbf{0.4849}}} & \multicolumn{1}{r|}{\cellcolor[gray]{0.85}{\color[HTML]{FE0000} \textbf{0.6592}}} & 0.4269                                                         & \cellcolor[gray]{0.85}{\color[HTML]{FE0000} \textbf{0.4849}}              \\ 
\textbf{LMI-AL(Coreset)}           & \multicolumn{1}{r|}{0.4021}                                                         & \multicolumn{1}{r|}{\cellcolor[gray]{0.85}\textbf{0.5165}}                        & 0.4339                                                         & \multicolumn{1}{r|}{0.4206}                                                         & \multicolumn{1}{r|}{\cellcolor[gray]{0.85}\textbf{0.5582}}                        & \cellcolor[gray]{0.85}\textbf{0.4636}                        & \multicolumn{1}{r|}{\cellcolor[gray]{0.85}\textbf{0.4655}}                        & \multicolumn{1}{r|}{0.5339}                                                         & 0.4656                                                         & {\cellcolor[gray]{0.85}\textbf{0.4655}}                                                                   \\ \hline
Random              & \multicolumn{1}{r|}{0.4165}                                                         & \multicolumn{1}{r|}{0.4792}                                                         & 0.5277                                                         & \multicolumn{1}{r|}{0.4226}                                                         & \multicolumn{1}{r|}{0.5969}                                                         & 0.4038                                                         & \multicolumn{1}{r|}{0.4241}                                                         & \multicolumn{1}{r|}{0.4362}                                                         & 0.4740                                                          & 0.4241                                                                     \\ 
Coreset             & \multicolumn{1}{r|}{0.4019}                                                         & \multicolumn{1}{r|}{0.5128}                                                         & 0.4680                                                         & \multicolumn{1}{r|}{0.4544}                                                         & \multicolumn{1}{r|}{0.5419}                                                         & 0.4590                                                         & \multicolumn{1}{r|}{0.3951}                                                         & \multicolumn{1}{r|}{0.4572}                                                         & 0.5014                                                         & 0.4544                                                                   \\ \hline
Percentage of $\mathcal{D}_L$ in SVUH              & \multicolumn{3}{c V{3}}{3\%}                                                                                                                                                                                                                   & \multicolumn{3}{c V{3}}{4\%}                                                                                                                                                                                                                   & \multicolumn{3}{c V{3}}{5\%}                                                                                                                                                                                                                   & \multicolumn{1}{l|}{\cellcolor[HTML]{FFFFFF}}                               \\ \cline{1-10}
Metrics                    & \multicolumn{1}{l|}{dice}                                                           & \multicolumn{1}{l|}{recall}                                                         & \multicolumn{1}{l V{3}}{precision}                                 & \multicolumn{1}{l|}{dice}                                                           & \multicolumn{1}{l|}{recall}                                                         & \multicolumn{1}{l V{3}}{precision}                                 & \multicolumn{1}{l|}{dice}                                                           & \multicolumn{1}{l|}{recall}                                                         & \multicolumn{1}{l V{3}}{precision}                                 & \multicolumn{1}{l|}{\multirow{-2}{*}{\cellcolor[HTML]{FFFFFF}Highest dice}} \\ \hline
LMI-AL(Random)                    & \multicolumn{1}{r|}{\cellcolor[gray]{0.85}{\color[HTML]{FE0000} \textbf{0.6272}}} & \multicolumn{1}{r|}{\cellcolor[gray]{0.85}{\color[HTML]{FE0000} \textbf{0.1849}}} & 0.4222                                                         & \multicolumn{1}{r|}{0.5776}                                                         & \multicolumn{1}{r|}{0.0814}                                                         & 0.4136                                                         & \multicolumn{1}{r|}{0.6152}                                                         & \multicolumn{1}{r|}{0.1660}                                                          & 0.4401                                                         & 0.6067                                                                      \\ 
LMI-AL(Margin)                     & \multicolumn{1}{r|}{0.6114}                                                         & \multicolumn{1}{r|}{\cellcolor[gray]{0.85}\textbf{0.1533}}                        & 0.3410                                                          & \multicolumn{1}{r|}{0.5744}                                                         & \multicolumn{1}{r|}{0.0746}                                                         & 0.4080                                                          & \multicolumn{1}{r|}{0.5839}                                                         & \multicolumn{1}{r|}{0.0996}                                                         & 0.3830                                                          & 0.5899                                                                      \\ 
LMI-AL(LeastConfidence)            & \multicolumn{1}{r|}{0.5572}                                                         & \multicolumn{1}{r|}{0.0508}                                                         & 0.3935                                                         & \multicolumn{1}{r|}{0.5968}                                                         & \multicolumn{1}{r|}{0.1239}                                                         & 0.4054                                                         & \multicolumn{1}{r|}{0.6111}                                                         & \multicolumn{1}{r|}{0.1469}                                                         & 0.4054                                                         & 0.5884                                                                      \\ 
LMI-AL(Entropy)                    & \multicolumn{1}{r|}{0.6099}                                                         & \multicolumn{1}{r|}{0.1464}                                                         & 0.3916                                                         & \multicolumn{1}{r|}{0.5996}                                                         & \multicolumn{1}{r|}{0.1204}                                                         & \cellcolor[gray]{0.85}{\color[HTML]{FE0000} \textbf{0.4344}} & \multicolumn{1}{r|}{0.6215}                                                         & \multicolumn{1}{r|}{0.1953}                                                         & \cellcolor[gray]{0.85}{\color[HTML]{FE0000} \textbf{0.4700}}   & 0.6103                                                                      \\ 
LMI-AL(BALD)                       & \multicolumn{1}{r|}{0.6073}                                                         & \multicolumn{1}{r|}{0.1261}                                                         & \cellcolor[gray]{0.85}\textbf{0.4350}                         & \multicolumn{1}{r|}{0.6275}                                                         & \multicolumn{1}{r|}{0.1913}                        & \cellcolor[gray]{0.85}\textbf{0.4203}                        & \multicolumn{1}{r|}{0.6163}                                                         & \multicolumn{1}{r|}{0.1525}                                                         & 0.4268                                                         & 0.6170                                                                      \\ 
LMI-AL(Coreset)                    & \multicolumn{1}{r|}{0.5927}                                                         & \multicolumn{1}{r|}{0.1028}                                                         & 0.4256                                                         & \multicolumn{1}{r|}{0.6236}                        & \multicolumn{1}{r|}{0.1817}                                                         & 0.4026                                                         & \multicolumn{1}{r|}{\cellcolor[gray]{0.85}{\color[HTML]{FE0000} \textbf{0.6322}}} & \multicolumn{1}{r|}{\cellcolor[gray]{0.85}\textbf{0.1962}}                        & \cellcolor[gray]{0.85}\textbf{0.4412}                        & 0.6162                                                                      \\ 
\textbf{LMI-AL(ClusterMargin)}     & \multicolumn{1}{r|}{\cellcolor[gray]{0.85}\textbf{0.6180}}                         & \multicolumn{1}{r|}{0.1484}                                                         & 0.4346                                                         & \multicolumn{1}{r|}{\cellcolor[gray]{0.85}{\color[HTML]{FE0000} \textbf{0.6574}}}                                                         & \multicolumn{1}{r|}{\cellcolor[gray]{0.85}{\color[HTML]{FE0000} \textbf{0.3003}}}                                                         & 0.4083                                                          & \multicolumn{1}{r|}{\cellcolor[gray]{0.85}\textbf{0.6288}}                        & \multicolumn{1}{r|}{\cellcolor[gray]{0.85}{\color[HTML]{FE0000} \textbf{0.2552}}} & 0.3969                                                         & \cellcolor[gray]{0.85}{\color[HTML]{FE0000}\textbf{0.6347}}                                     \\ 
\textbf{LMI-AL(Hybrid)}            & \multicolumn{1}{r|}{0.6125}                                                         & \multicolumn{1}{r|}{0.1438}                                                         & \cellcolor[gray]{0.85}{\color[HTML]{FE0000} \textbf{0.4395}} & \multicolumn{1}{r|}{\cellcolor[gray]{0.85}\textbf{0.6512}} & \multicolumn{1}{r|}{\cellcolor[gray]{0.85}{\textbf{0.2207}}} & 0.4506                                                         & \multicolumn{1}{r|}{0.6094}                                                         & \multicolumn{1}{r|}{0.1293}                                                         & 0.4406                                                         & \cellcolor[gray]{0.85}\textbf{0.6244}              \\ \hline
ClusterMargin & \multicolumn{1}{r|}{0.6016}                                                         & \multicolumn{1}{r|}{0.1815}                                                         & 0.3263                                                         & \multicolumn{1}{r|}{0.6485}                                                         & \multicolumn{1}{r|}{0.2483}                                                         & 0.4337                                                         & \multicolumn{1}{r|}{0.6149}                                                         & \multicolumn{1}{r|}{0.2183}                                                         & 0.3253                                                         & 0.6217                                                                      \\ 
Hybrid        & \multicolumn{1}{r|}{0.5927}                                                         & \multicolumn{1}{r|}{0.1358}                                                         & 0.3618                                                         & \multicolumn{1}{r|}{0.6229}                                                         & \multicolumn{1}{r|}{0.1552}                                                         & 0.4699                                                         & \multicolumn{1}{r|}{0.6080}                                                          & \multicolumn{1}{r|}{0.1746}                                                         & 0.3425                                                         & 0.6079                                                                      \\ \hline
\end{tabular}}
\end{center}
\label{tab:results}
\end{table*}

Table \ref{tab:results} presents the experimental results for the MSSEG-2 and SVUH datasets with varying percentages of labeled slice pairs. LMI-AL achieves comparable performance to a DL trained on a fully labeled dataset with only 8\% (e.g., LMI-AL(Random)) of labeled slice pairs for MSSEG-2 and 5\% (e.g., LMI-AL(Coreset) and LMI-AL(ClusterMargin)) for SVUH. The results demonstrate that even with a small percentage of labeled data, LMI-AL strategies can achieve performance close to that of fully supervised models. The Highest dice scores achieved in the process show that all AL methods perform similarly (dice values within 0.06 of the best value on MSSEG-2 and 0.03 on SVUH dataset) except for Least Confidence and Margin. This could be because medical imaging datasets often suffer from class imbalance, with a large number of negative (no change) slice pairs compared to positive (change) slice pairs. Both of these methods use simple approaches to measure uncertainty and do not inherently address class imbalance, which may result in sub-optimal selection of positive samples for labeling.

To further evaluate the effectiveness of the proposed method, we compared the two best-performing DAL strategies for each dataset (e.g., Random and Coreset for MSSEG-2, ClusterMargin and Hybrid for SVUH) with and without integration into the LMI-AL framework. This comparison helps assess the actual improvements achieved by LMI-AL in change detection for longitudinal images. On MSSEG-2, both Random and Coreset strategies, when integrated with LMI-AL, show improved performance across different percentages of labeled slice pairs, with the highest increase observed at 8\% labeled slice pairs (Random shows a 0.06 increase and Coreset shows a 0.07 increase in dice). Similarly, in SVUH, ClusterMargin and Hybrid also show performance gains with LMI-AL across different portions of labeled data.

% To further evaluate the effectiveness of the proposed method, we compared the two best-performing AL strategies for each dataset (e.g., Random and Coreset for MSSEG-2, ClusterMargin and Hybrid for SVUH) with and without integration into the LMI-AL framework. This comparison helps assess the actual improvements achieved by LMI-AL in change detection for longitudinal images. On MSSEG-2, integrating Random and Coreset with LMI-AL enhances performance, particularly at 8\% labeled slice pairs, where dice scores increase by 0.06 and 0.07, respectively. Similarly, in SVUH, ClusterMargin and Hybrid also show performance gains with LMI-AL across different portions of labeled data.

\subsection{Analysis and Discussion}
Although supervised learning results across datasets are inconsistent (shown in Table~\ref{tab1}), the outcomes in the DAL settings reveal similar trends (shown in Table \ref{tab:results} ). Here in this section, we provide corresponding analysis.

% 从model角度 学习惰性 学习倾向性 理论分析

\subsubsection{Minimal Sample Sufficiency}
\begin{figure}[htbp] % 使用 figure* 以使图跨越双栏
\centering
\begin{subfigure}[b]{0.23\textwidth} % 将宽度微调为更接近文本宽度的一半
    \includegraphics[width=\textwidth]{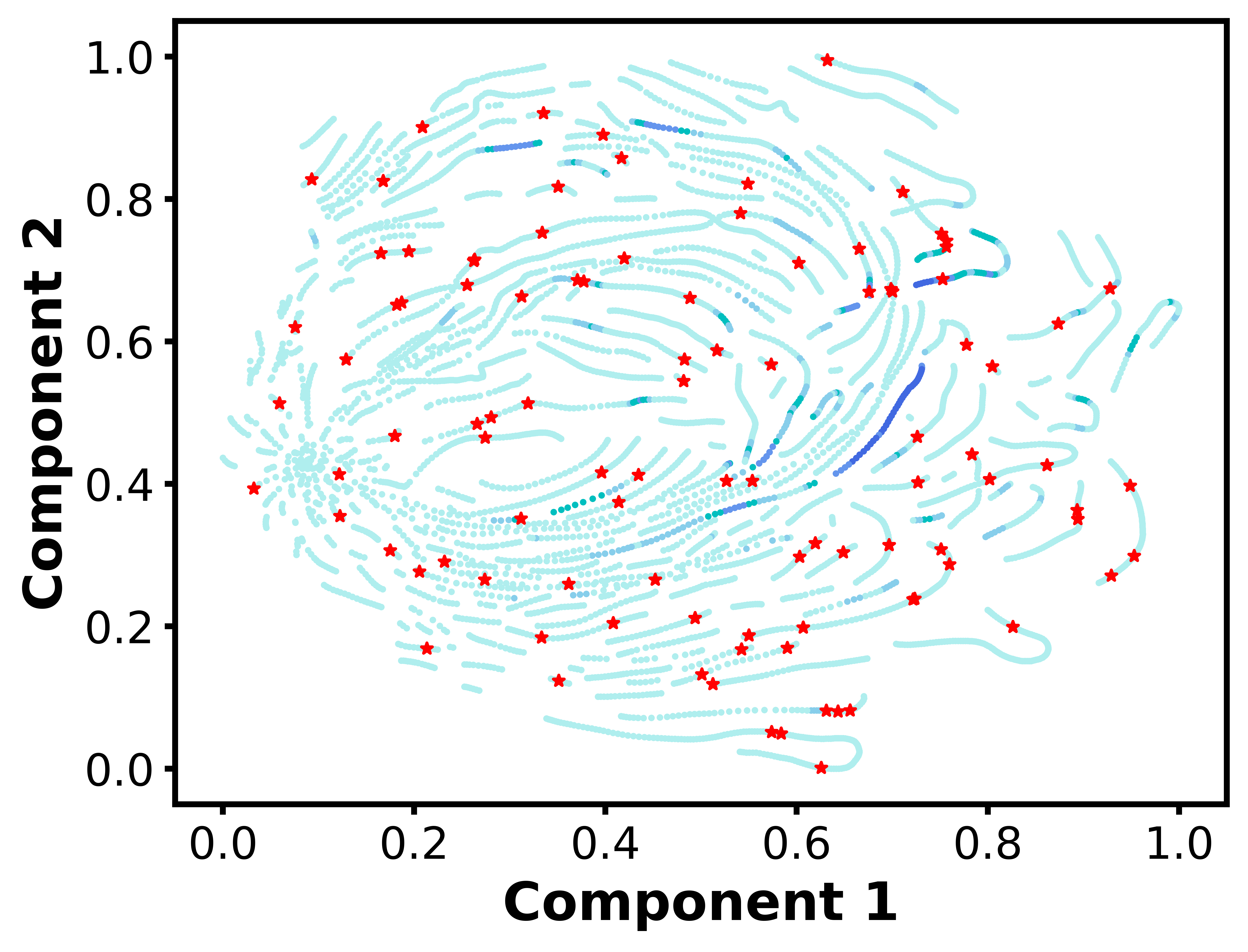}
    \caption{Random selection}
    \label{fig:random}
\end{subfigure}
\begin{subfigure}[b]{0.23\textwidth} % 去掉hfill，并调整子图宽度来减小间隙
    \includegraphics[width=\textwidth]{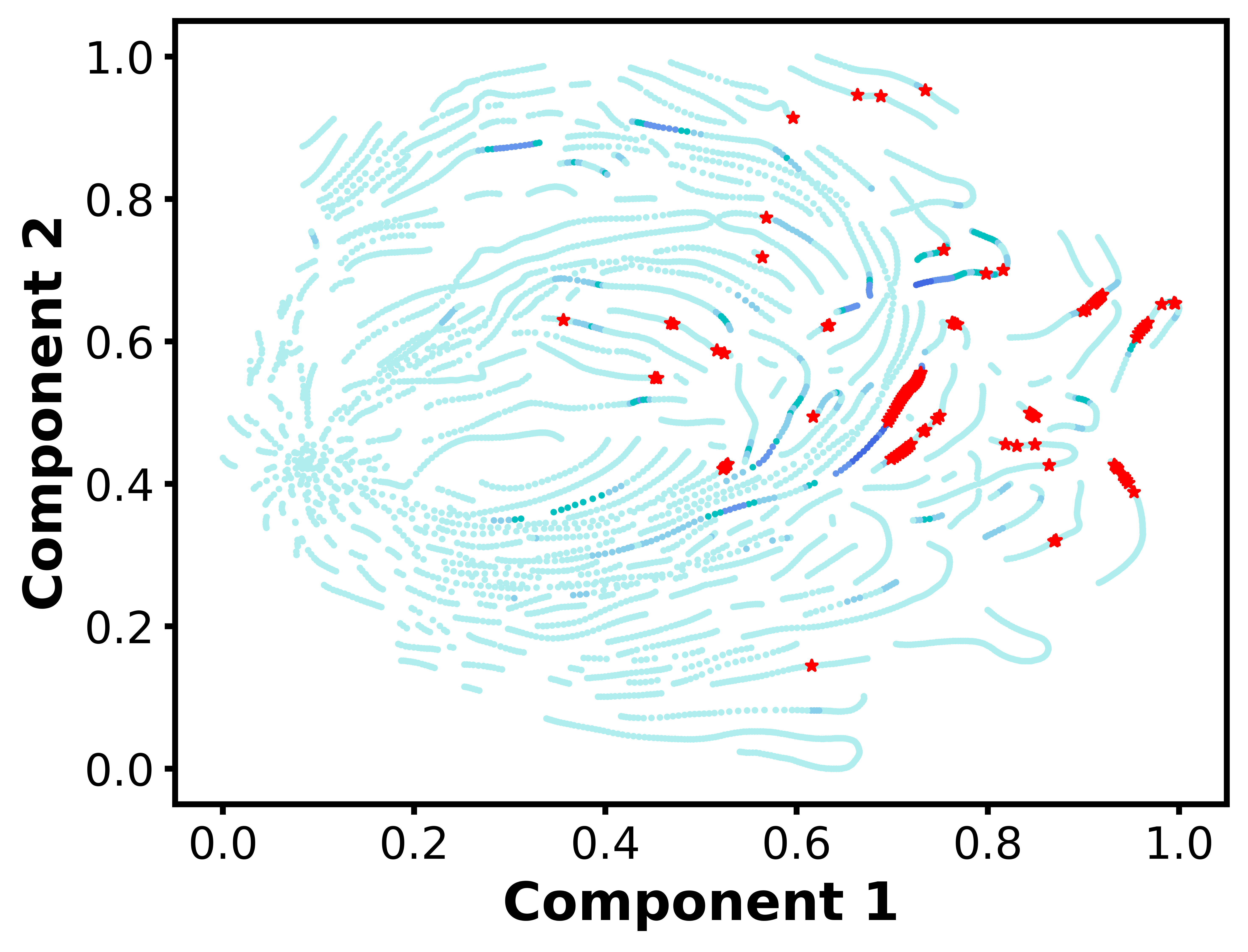}
    \caption{Entropy-based selection}
    \label{fig:entropy}
\end{subfigure}
\vspace{0pt} % 控制子图之间的垂直距离
\begin{subfigure}[b]{0.23\textwidth} % 相同宽度设置以确保对齐
    \includegraphics[width=\textwidth]{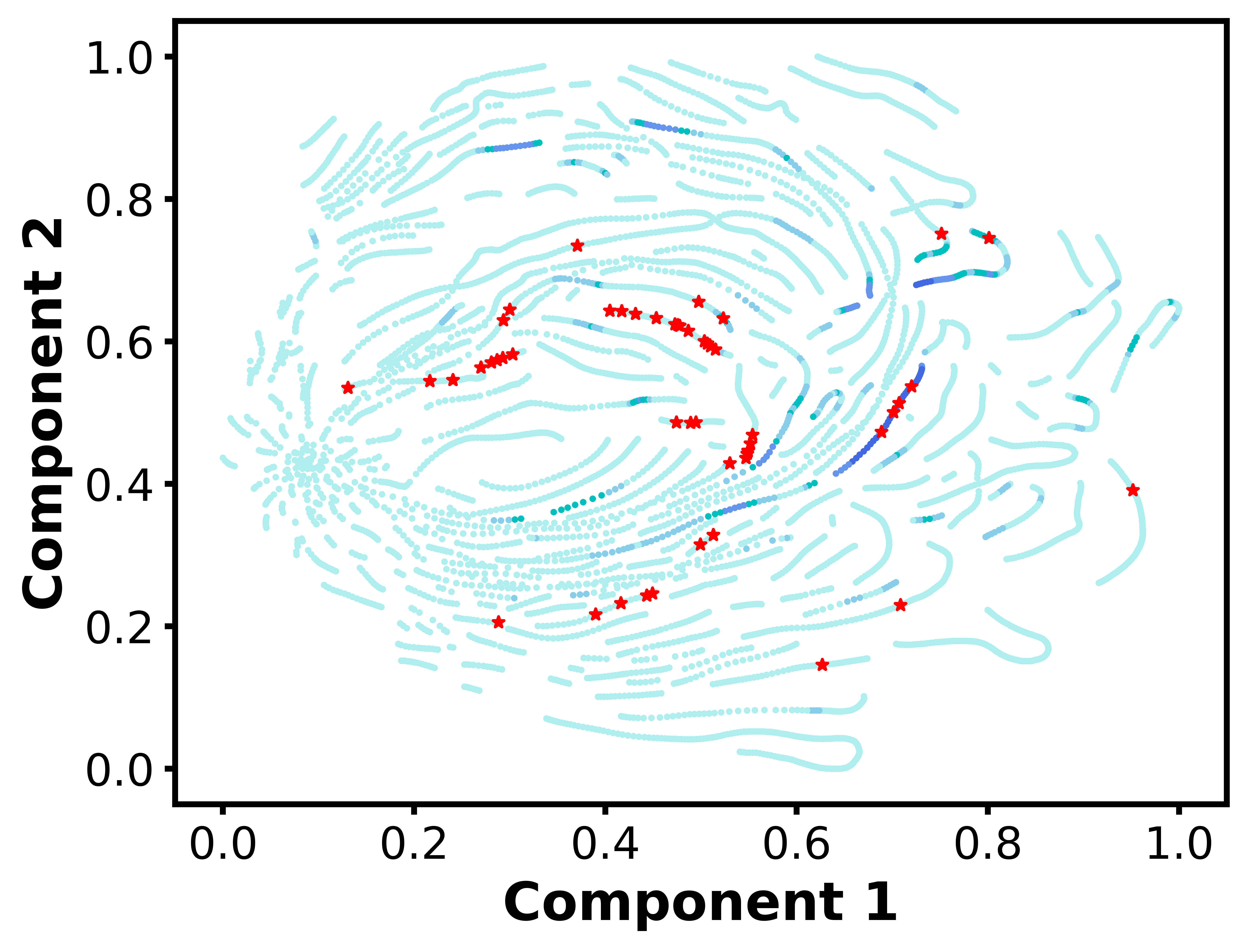}
    \caption{Coreset selection}
    \label{fig:coreset}
\end{subfigure}
\begin{subfigure}[b]{0.23\textwidth}
    \includegraphics[width=\textwidth]{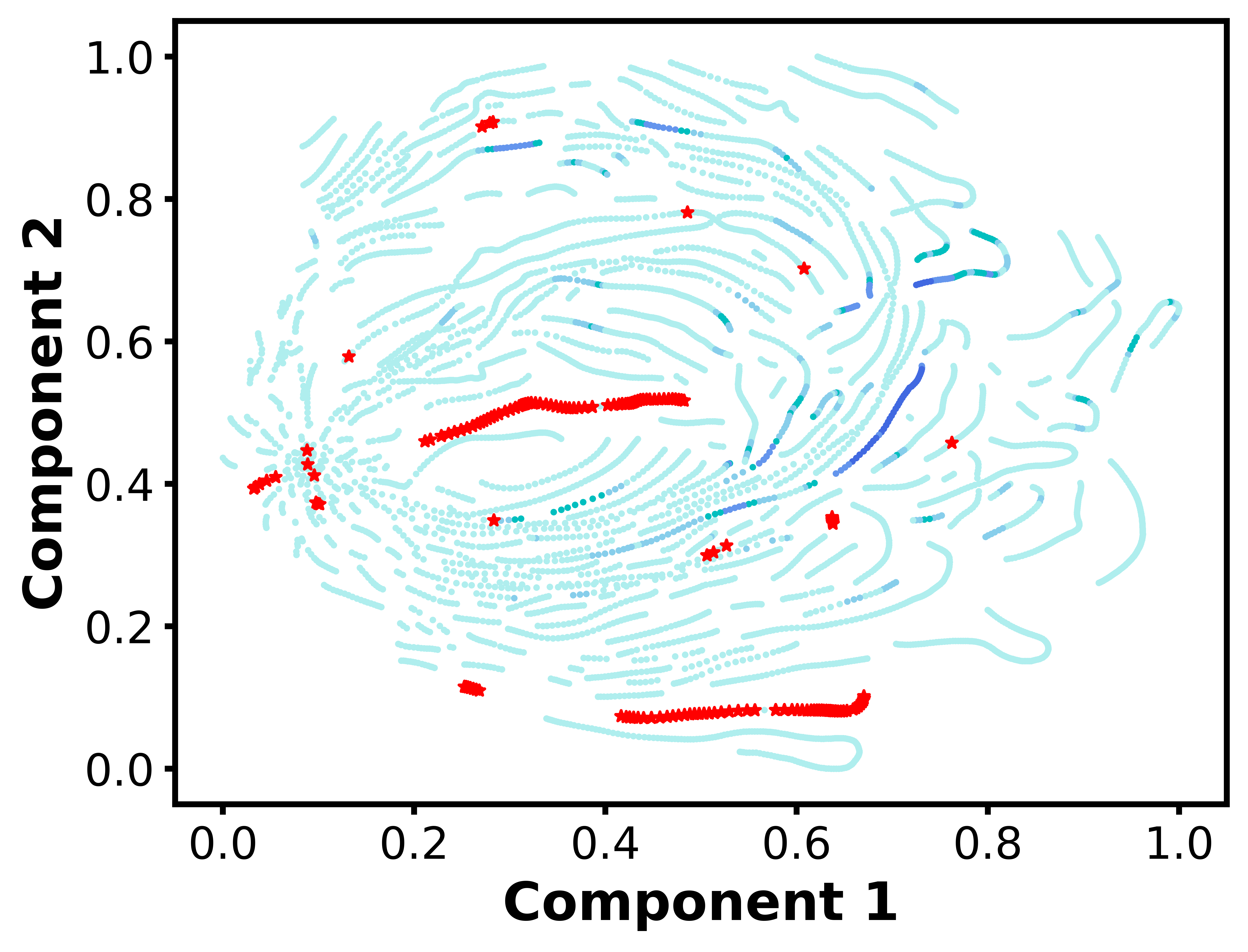}
    \caption{ClusterMargin method}
    \label{fig:hybrid}
\end{subfigure}
\caption{Visualizations of sample selections by different active learning methods. Red points indicate the selected samples up to the point where performance matches the fully supervised baseline, while blue points represent all available unlabeled images in the pool.}
\label{fig2}
\end{figure}

Both datasets demonstrate that a small portion of labeled data can achieve comparable performance compared to the supervised learning baseline (less than 8\% of the data was required for the MSSEG-2 dataset and less than 5\% for the SVUH dataset). \textbf{This efficiency may be attributed to the high redundancy inherent in the datasets.} To further demonstrate this, we visualized the distribution of slice pair selections made by the top-performing uncertainty- (LMI-AL(Entropy)), diversity- (LMI-AL(Random) and LMI-AL(Coreset)), and hybrid-based AL (LMI-AL(ClusterMargin)) methods on the MSSEG-2 dataset. Figure \ref{fig2} shows representative results, where the number of target-containing pixels in each slice is depicted in shades of blue, with darker shades indicating more targets. Due to the different principles behind each query strategy, the distribution of selected slice pairs varies greatly. However, it is interesting to note that their performance does not show significant differences (see Table \ref{tab:results}). We suppose this is due to the redundancy of longitudinal medical datasets (e.g., many slices with new lesions that are very similar in shape, size, and location), demonstrating the importance of DAL in this setting. 

% \subsubsection{Ablation Study}
% To validate the impact of incorporating the differences between different time slots (t1-t0's difference) in the AL framework, we selected the top 2 methods from each dataset (Random and Coreset for MSSEG2, and ClusterMargin and Hybrid Sampling for the private dataset, as shown in Tab.\ref{tab4} and Tab.\ref{tab5}). Specifically, given a patient with three time slot (t0, t1, t2), to demonstrate the difference between using all the combination slices (t0, t1), (t0, t2), and (t1, t2) with total 29638 pairs or consecutive slices (t0, t1) and (t1, t2) (22569 pairs), we conducted this experiment in Tab.\ref{tab5} as well. Our findings indicate that, in most cases, adding the difference leads to superior results, particularly notable in achieving the highest performance consistently across both datasets. This is because it guides the model to focus on crucial areas of change, thereby aiding the change detection task. 

\subsubsection{Suitability of Diversity-based and Hybrid Methods}
As shown in table \ref{tab:results}, the diversity and hybrid query strategies perform better in both two dataset. \textbf{The success of these methods likely relates to the limited budget of labeled samples necessary for training.} Since the required number of samples is relatively low, the models benefit significantly from learning a general distribution of the data \cite{ma2024adaptive}. Hybrid-based method that dynamically adjust between diversity and uncertainty-based methods could potentially harness the strengths of both strategies. While uncertainty-based methods typically become more relevant with increased labeling budgets, \textbf{our analysis suggests that in the context of MS longitudinal datasets, extensive budgets may not be necessary. Therefore, diversity-based and hybrid methods should be applied when a new longitudinal dataset is given. } 

\subsubsection{Comparison between Active learning and Fully supervised learning} 
In the MSSEG2 dataset, AL results in table \ref{tab:results} show that recall is comparable to that of supervised learning, while precision is on average higher. This can be attributed to the limited sample amount of MSSEG2, where supervised learning needs the full dataset to yield better performance. Despite achieving higher precision compared to using only target-containing samples, the dataset remains highly imbalanced (the area of target is quite small compared to background). However, \textbf{DAL methods can mitigate the imbalance problem}. For example, Coreset, as shown in Figure~\ref{fig2}(c), even focusing on diversity-based selection, the sample selection tends to focus on areas with more target pixels, reducing the level of imbalance and resulting in a precision of 0.4656, compared to 0.4467 in supervised learning. 

On the other hand, using only target-containing samples in supervised learning demonstrates higher recall, reaching 0.6674 compared to 0.6097. AL achieves similar results using only 8\% of the samples, where Hybrid and Random sampling reach 0.6959 and 0.6592, respectively. \textbf{This suggests that even a small number of blank and target-containing samples can perform comparably to using all target samples, likely due to the broader distribution of blank samples.}

The SVUH dataset yields similar conclusions. Even when using all samples with target for supervised learning, AL results achieve comparable or even superior outcomes, significantly surpassing the results obtained when using all slices (0.5907, 0.1049, 0.4109). This indicates that the SVUH data contains a substantial amount of redundancy, and that blank samples are beneficial to the model's learning process. 

\section{Conclusion}
\label{Conclusion}
In this paper, we investigate an important and previously unexplored problem: training a model for change detection in longitudinal medical imaging using limited labeled data. We propose a novel framework LMI-AL that adapts active learning for change detection in longitudinal imaging. The experimental results show that using less than 8\% of labeled data allows the deep learning model to achieve performance similar to training with a fully labeled dataset, which indicates that labeling all the data in longitudinal medical imaging is unnecessary. We also provide detailed analysis and give guidance for future related research. Considering the difficulties and high cost of labeling longitudinal medical images, our approach offers a practical solution to efficiently train models with minimal labeling effort, potentially reducing both time and resources in clinical applications. Future work could focus on analyzing the spatial distribution of the selected slices through visualization, revealing which regions are more frequently sampled. This may provide further insights into the model's learning behavior and patterns.

\section{Acknowledgment}
This publication has emanated from research conducted with the financial support of Science Foundation Ireland under Grant number [12/RC/2289\_P2]. 

\bibliographystyle{IEEEtran}
\bibliography{IEEEexample}
\end{document}